\documentclass[journal]{IEEEtran}

\usepackage{xcolor,soul,framed} 

\colorlet{shadecolor}{yellow}
\usepackage[pdftex]{graphicx}
\graphicspath{{../pdf/}{../jpeg/}}
\DeclareGraphicsExtensions{.pdf,.jpeg,.png}

\usepackage[cmex10]{amsmath}
\usepackage{array}
\usepackage{mdwmath}
\usepackage{mdwtab}
\usepackage{eqparbox}
\usepackage{url}
\usepackage{bbding}
\usepackage{multirow,subfigure,pifont,hyperref}
\usepackage[inkscapelatex=false]{svg}
\usepackage{booktabs} 
\usepackage{amssymb}
\newcommand{\tabincell}[2]{\begin{tabular}{@{}#1@{}}#2\end{tabular}}
\usepackage{color}
\usepackage{bm}
\usepackage{yhmath}
\usepackage{pifont}
\usepackage{bbding}
\usepackage{array}
\usepackage{nicematrix}
\usepackage{fontawesome}

\begin{document}
\bstctlcite{IEEEexample:BSTcontrol}
    \title{Progressive Domain Adaptation for Thermal Infrared Object Tracking}
  \author{Qiao~Li,
      Kanlun Tan,
      Qiao~Liu,
      ~Di~Yuan,
      ~Xin~Li,
      Yunpeng~Liu}



\maketitle

\begin{abstract}
Due to the lack of large-scale labeled Thermal InfraRed (TIR) training datasets, most existing TIR trackers are trained directly on RGB datasets. 
  However, tracking methods trained on RGB datasets suffer a significant drop-off in TIR data due to the domain shift issue. To this end, in this work, we propose a Progressive Domain Adaptation framework for TIR Tracking (PDAT), which transfers useful knowledge learned from RGB tracking to TIR tracking. The framework makes full use of large-scale labeled RGB datasets without requiring time-consuming and labor-intensive labeling of large-scale TIR data. Specifically, we first propose an adversarial-based global domain adaptation module to reduce domain gap on the feature level coarsely. Second, we design a clustering-based subdomain adaptation method to further align the feature distributions of the RGB and TIR datasets finely. These two domain adaptation modules gradually eliminate the discrepancy between the two domains, and thus learn domain-invariant fine-grained features through progressive training. Additionally, we collect a large-scale TIR dataset with over 1.48 million unlabeled TIR images for training the proposed domain adaptation framework. Experimental results on five TIR tracking benchmarks show that the proposed method gains a nearly  6\% success rate, demonstrating its effectiveness. 
\end{abstract}


%
\IEEEpeerreviewmaketitle

\section{Introduction}
\label{introduction}

%
\IEEEPARstart{S}{ince} TIR imaging has the advantage of not being affected by illumination variation, visual object tracking technology based on TIR images begin to receive extensive research in recent years. It can be widely used in nighttime autonomous driving, video surveillance, and missile tracking etc. Compared with RGB images, TIR images have low resolution, lack of rich texture details, and low signal-to-noise ratio. These characteristics make TIR tracking more difficult than RGB tracking, and it is easily affected by background clutters or similar distractors and causes drift.

\begin{figure}[t]
\centering
\includegraphics[width=8.5cm]{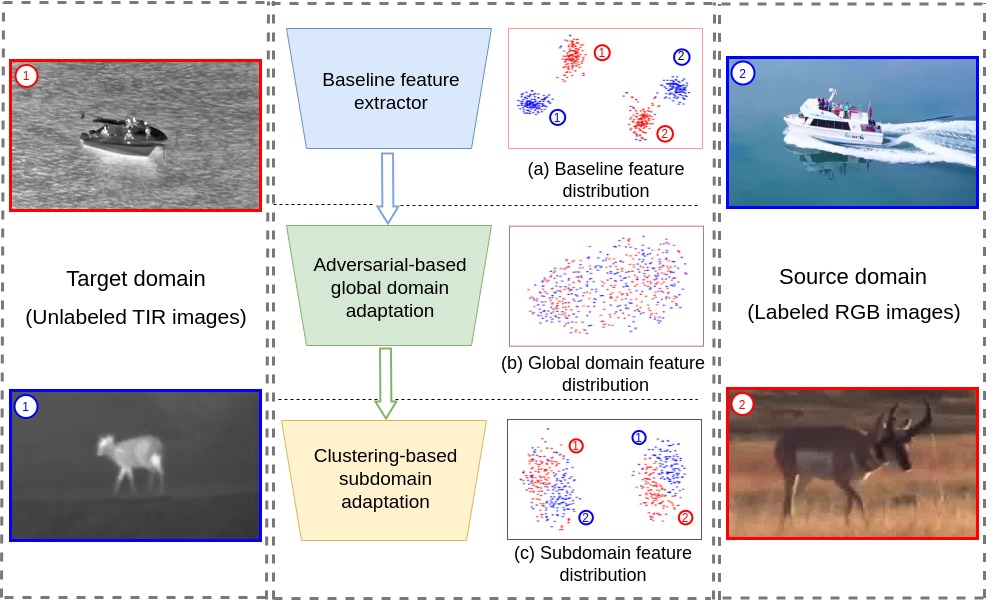}
\caption{Visualization of the feature distribution of the baseline and the proposed progressive domain adaptation model based on t-SNE. Numbers \ding{172} and \ding{173} represent the TIR and RGB domain samples, respectively. It shows that the distance between the feature distributions of similar samples in different domains extracted by the baseline feature extractor is very large. While after using our global domain adaptation and subdomain adaptation modules, the obtained feature distributions are gradually narrowed.}
\label{feature_vis}
\end{figure}

To overcome these challenges, early TIR tracking methods mainly focus on how to extract the more discriminative features of TIR targets~\cite{review}, including intensity histogram~\cite{venkataraman2012adaptive}, improved gradient direction histogram \cite{yu2017dense}, distribution field \cite{berg2016channel}, and multi-feature fusion \cite{gao2016infrared,asha2017robust,xu2021hierarchical}. In recent years, some works attempt to utilize the powerful representation capabilities of deep features to improve TIR tracking. For example, MCFTS~\cite{MCFTS} and LMSCO \cite{LMSCO} use pre-trained VGGNet to extract deep feature of TIR target. HSSNet \cite{li2019hierarchical}, MLSSNet \cite{MLSSNet}, and DFG \cite{DFG} train a deep matching network on large-scale RGB dataset and then used for TIR tracking directly. ECO-stir \cite{zhang2018synthetic} trains a deep correlation filter network on synthetic TIR images which translated from RGB images. Similar, MMNet~\cite{MMNet} first trains the matching network on large-scale RGB dataset and then fine-tuning on a small-scale labeled TIR dataset. Due to the lack of large-scale labeled TIR datasets, it is difficult to train a high-performance TIR tracker from scratch. This is why most existing TIR tracking methods are only trained on RGB datasets. However, due to different imaging principles, there are significant differences between TIR images and RGB images. As a result, tracking methods trained on RGB images have a significant drop-off on TIR data. We attribute the main reason to the domain shift problem between the TIR and RGB datasets. 
As shown in Figure~\ref{feature_vis}(a), we can see that there is a significant distribution discrepancy between the similar class samples in the TIR and RGB domains when using the baseline feature extractor trained on RGB dataset. This illustrates that deep features learned on RGB datasets cannot be well generalized to TIR tracking.

\begin{figure*}[t]
\centering
\scalebox{0.55}{
\includegraphics{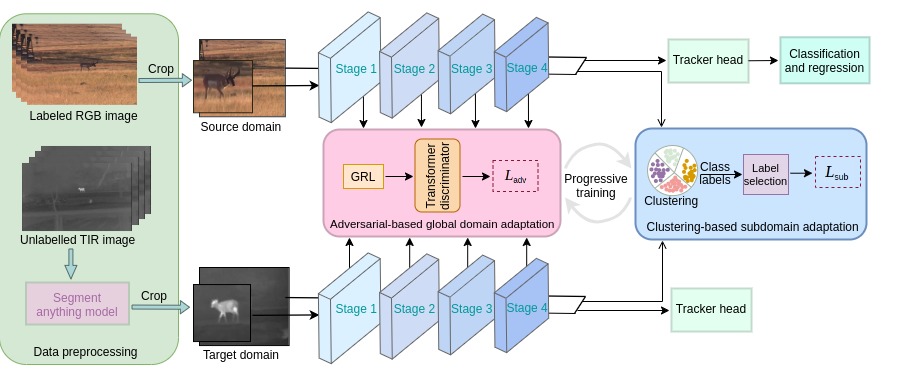}
}
\caption{Proposed progressive domain adaptation TIR tracking framework (PDAT) which mainly consists of three parts: Segment Anything Model (SAM) based data preprocessing, Adversarial-based Global Domain Adaptation (AGDA), and Clustering-based SubDomain Adaptation (CSDA). SAM is used to generate a large number of pseudo-labeled TIR training data likes source samples. AGDA aligns global domain coarsely, while CSDA further aligns subdomains finely. Stage 1 to Stage 4 denote the feature extraction block of the backbone. 
}
\label{Overall structure}
\end{figure*}

To address the domain shift and performance degradation challenges, a potential direct solution is to collect and annotate a large number of trainable TIR samples. However, this is time-consuming and expensive. 
Instead of annotating a large-scale TIR dataset, in this paper, we propose a progressive domain adaptation framework for TIR tracking, called PDAT. 
This framework can make full use of large-scale labeled RGB dataset to learn tracking universal priors, and then transfer to TIR tracking only using unlabeled TIR data.
Specifically, we propose a global domain adaptation and a subdomain adaptation module to progressive align the feature distributions of the TIR and RGB datasets, which can learn domain-invariant features for TIR tracking.
For the global domain adaptation module, we use an adversarial learning way to roughly narrow the feature distribution of the overall TIR and RGB domains.
However, simply aligning the global domain feature distribution is not sufficient for tracking task that require fine-grained features because the tracking task needs to distinguish similar targets.
Therefore, we propose a clustering-based subdomain adaptation method to align features in the subdomain with similar category for getting more fine-grained feature transfer.
The proposed global domain adaptation and subdomain adaptation gradually align the feature distributions of the TIR and RGB domains in a progressive training manner. Global domain adaptation can provide a good initialization for subdomain adaptation, while subdomain adaptation can accelerate global domain alignment.
Figure \ref{feature_vis} shows that our method can effectively align the feature distribution of the RGB and TIR datasets to achieve fine-grained feature transfer.
Extensive experimental results on five TIR tracking benchmarks, including LSOTB-TIR100, LSOTB-TIR120, PTB-TIR, VTUAV and VOT-TIR2017, demonstrate that the proposed method achieves favorable performance.
The main contributions are summarized as follows:
\noindent
\begin{itemize}
     \item We propose a progressive domain adaptation framework for TIR tracking. To the best of our knowledge, this is the first domain adaptation framework specifically designed for TIR tracking.
     \item We propose an adversarial-based global domain adaptation and a clustering-based subdomain adaptation modules. These two domain adaptation modules progressively align the feature distributions of the two domains to achieve fine-grained feature transfer.
      \item We collect a large-scale unlabeled TIR dataset and generate a pseudo-label training set through the SAM model for training the proposed framework.
     \item We conduct extensive experiments on five TIR tracking benchmarks and the results demonstrate that our method achieves favorable performance. 
\end{itemize}

\section{Related Work}
\label{relatedwork}
\subsection{Deep TIR trackers} 
According to different training strategies, deep TIR trackers can be roughly divided into three categories: pre-trained, training from scratch, and fine-tuning. 
Pre-trained based deep TIR trackers usually use a pre-trained feature network learned from RGB datasets to extract the deep feature of TIR targets, and then combine an existing tracking framework. For example, MCFTS~\cite{MCFTS} uses pre-trained VGGNet~\cite{VGGNet} to get multiple convolution features of the TIR target and then combine with a correlation filter to form an ensemble-based TIR tracker. Similar, LMSCO~\cite{LMSCO} use pre-trained VGGNet and MotionNet~\cite{MotionNet} to extract appearance and motion feature of TIR targets and then integrates them into a structural support vector machine.
Training for scratch based deep TIR trackers usually train a convolution neural network on large-scale RGB datasets and then used for TIR tracking directly. For instances, both HSSNet~\cite{li2019hierarchical}, MLSSNet~\cite{MLSSNet} and DFG~\cite{DFG} train a deep matching network on several large-scale RGB tracking datasets and then used for TIR tracking. DSST-TIR~\cite{DSST-TIR} trains a deep classification network on a small-scale labeled TIR dataset. ECO-stir~\cite{zhang2018synthetic} trains a deep correlation filter network on the synthetic TIR dataset which generated from large-scale RGB tracking datasets.
Fine-tuning based deep TIR trackers first train a deep network from RGB datasets and then fine-tuning on a small-scale TIR dataset to adapt the feature to the TIR tracking. For example, MMNet~\cite{MMNet} and TransT-FTIR~\cite{liu2023lsotb} first train a deep Siamese network on large-scale RGB tracking datasets and then fine-tuning on a limited labeled TIR tracking dataset. 
Although these methods use the powerful representation capabilities of deep features to improve the performance of TIR trackers, due to the lack of large-scale labeled TIR training data, most of these methods are still trained on RGB dataset, making it difficult to effectively generalize to TIR tracking and get excellent performance. To this end, we propose a domain adaptation framework that can fully transfer the knowledge learned on large-scale RGB dataset to TIR tracking.

\subsection{Domain adaptation}
The goal of domain adaptation is to reduce the distribution discrepancy between the source domain (usually labeled data) and the target domain (usually unlabeled data) to enhance the model's generalization ability. 
Because domain adaptation theory has strong practical application capabilities, it has been widely studied in the computer vision field in recent years.
For example, \cite{chen2018domain} introduces a domain-adaptive Faster RCNN framework, resolving the domain-adaptive object detection problem for the first time. They utilize the H-distance to measure the differences between two domains and align feature distributions through adversarial training. 
\cite{ma2022i2f} proposes a global luminance alignment and a texture alignment module, aligning both at the image and feature levels for boosting performance of semantic segmentation. 
\cite{ye2022unsupervised} combines Transformer with adversarial network to eliminate distribution discrepancy between day and night images, and then trains an adaptive tracker for nighttime unmanned aerial vehicles. 
Inspired by this work, we propose a progressive domain adaptation framework to transfer the useful knowledge learned from large-scale RGB dataset to TIR tracking.

\section{Proposed Method}
\label{method}
As shown in Figure~\ref{Overall structure}, the proposed method mainly contains three core components: segment anything model based data preprocessing, adverisarial-based global domain adaptation, and clustering-based subdomain adaptation. 
In the following, we will describe these three parts in detail.

\subsection{TIR dataset and preprocessing }
To train the proposed domain adaptation TIR tracking framework, we construct a large-scale unlabeled TIR training dataset. 
We first collect a large number of TIR videos from Internet, and then convert them to the TIR image sequences manually.
Since the data sources come from various platforms on the Internet, the quality of the data is uneven, but this also ensures that the dataset has good diversity.
As shown in Table~\ref{Dataset comp}, the collected TIR dataset contains $2549$ image sequences and over $1.48$ million frames, making this dataset the largest currently available in TIR tracking.

\begin{table}[ht]
    \centering
     \caption{Statistical comparison between the collected unlabeled TIR dataset and existing TIR and RGB tracking datasets.}
    \scalebox{0.72}{
    \fontsize{9}{12}\selectfont
    \begin{tabular}{l | l r r r r r c}
        \toprule
           &&Num. of  & Max & Min &Mean &Total &Training\\
          &Benchmarks&sequence &frame&frame&frame&frame&dataset\\
          \cline{2-8}
          &OSU  & 6 & 2,031 & 601   &1,424  &8K & \raisebox{-0.5ex}{\XSolidBrush}  \\ 
      
        &PDT-ATV       & 8 & 775   & 77   &486   &4K & \raisebox{-0.5ex}{\XSolidBrush}\\ 
        
         &BU-TIV   & 16 & 26,760 &150 &3,750 &60K & \raisebox{-0.5ex}{\XSolidBrush}\\ 
       
         &LTIR   & 20 & 1,451 & 71 &563 &11K & \raisebox{-0.5ex}{\XSolidBrush}\\ 
        
       TIR &VOT-TIR   & 25 & 1,451 & 71 &555 &14K & \raisebox{-0.5ex}{\XSolidBrush}\\ 
     
        Dataset&PTB-TIR   & 60 &1,451 & 50 &502 &30K & \raisebox{-0.5ex}{\XSolidBrush}\\ 
        
        &RGB-T   & 234 & 4,000 &45  &500 &117K & \raisebox{-0.5ex}{\XSolidBrush}\\ 
        
        &LSOTB-TIR(T) & 1,305 & 3,056 & 47 &415 &542K & \raisebox{-0.5ex}{\XSolidBrush}\\ 
        
        &LSOTB-TIR(ST) & 100 & 2,110 & 105 &643 &64K & \raisebox{-0.5ex}{\Checkmark}\\ 
        
        &LSOTB-TIR(LT) & 11 & 4,720 & 2,600 &3,449 &37K & \raisebox{-0.5ex}{\XSolidBrush}\\ 
        
        & TIR-nolabel(\textbf{Ours})  & \textbf{2,549} & \textbf{13,512} & \textbf{30} &\textbf{580} & \textbf{1.48M} & \raisebox{-0.5ex}{\Checkmark}\\ 
        \hline
        \hline
        &OTB15 & 100 & 3,872 & 71 &590 &59K & \raisebox{-0.5ex}{\XSolidBrush}\\ 
     
        &VOT17 & 60 & 1,500 & 41 &356 &21K & \raisebox{-0.5ex}{\XSolidBrush}\\ 
       
       RGB &UAV123 & 123 & 3,085 & 109 &915 &113K & \raisebox{-0.5ex}{\XSolidBrush}\\ 
        
        Dataset&LaSOT & 1,400 & 11,397 & 1,000 &2,506 &3.52M & \raisebox{-0.5ex}{\Checkmark}\\ 
       
         &GOT-10K   & 10,000 & - & - &150 &1.5M & \raisebox{-0.5ex}{\Checkmark}\\ 
        \bottomrule
    \end{tabular}
    }
    \label{Dataset comp}
\end{table}

\begin{figure*}[t]
\centering
\scalebox{0.4}{
\includegraphics{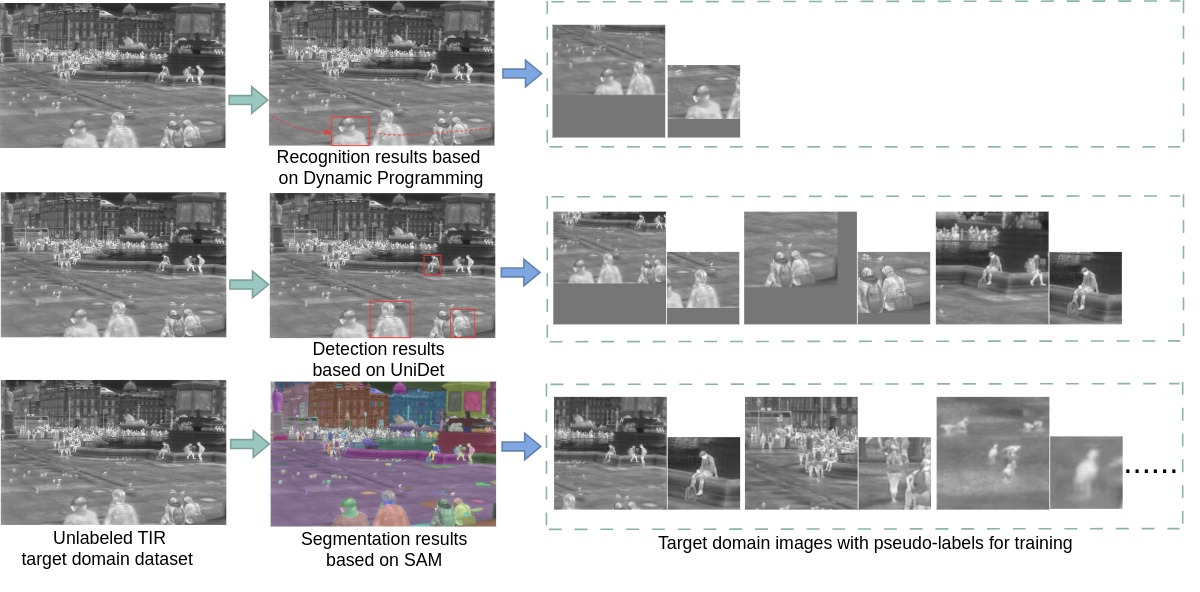}
}
\caption{Comparison of results obtained by three pseudo-label generation preprocessing methods. It can be seen that the SAM-based method obtains more sample pairs with higher diversity.
}
\label{SAM}
\end{figure*}

Since the proposed domain adaptation TIR tracking framework requires pairs of TIR samples for training, the collected TIR dataset does not contain label information. To this end, we utilize three kinds of methods, including Dynamic Programming-based recognition~\cite{ye2022unsupervised}, UniDet object detection model~\cite{UniDet}, and the segment anything model~\cite{kirillov2023segment} to generate pseudo label pairs. As shown in Figure~\ref{SAM}, we compare three methods for obtaining numerous potential target areas within TIR training data. SAM stands out as the most effective option, providing a greater diversity of training samples and yielding superior tracking results compared to other preprocessing methods. Initially, we plan to segment each frame of the TIR dataset using the SAM model. However, we discovered that this approach would require approximately six days to complete. To streamline this process, we adopt two key strategies. First, given the minimal differences between adjacent frames, we decided to segment every tenth frame. Second, we adjusted SAM's confidence threshold to higher levels, resulting in the generation of more precise yet fewer training image pairs from a single frame.These strategies significantly cut the preprocessing time from six days to approximately two days. Ultimately, we procure over nine million pairs of TIR training data, which exceeds the size of the initially collected TIR dataset by a factor of six. It is important to mention that we employ the SAM model for offline data preprocessing to generate pseudo labels. hence, the time expenditure for this step is not factored into the training phase. Consequently, our proposed domain adaptation method not only refines the baseline tracking framework but also maintains the tracking speed.

\subsection{Adversarial-based global domain adaptation}
Due to different imaging principles, the styles of TIR and RGB images are very different, which is the main reason for the domain shift problem. 
To effective transfer the tracking prior learned in RGB images to TIR tracking, in this paper, we hope to learn a domain-invariant feature which is not sensitive to the image style.

To achieve our objective, we introduce a global domain feature alignment network that employs an adversarial learning approach.  Specifically, we have designed a Transformer-based style discriminator, which is pivotal in enabling the feature model to discern the stylistic nuances of the input images.  The Transformer discriminator, through its self-attention mechanisms, adeptly captures the global dependencies present within an image.  Given the intrinsic link between an image's style and its holistic information, this global perspective is instrumental in differentiating the stylistic attributes of both the source and target domains.  Consequently, this enhanced discrimination leads to a more refined feature alignment, which is essential for effective domain adaptation.
The style discriminator consists of a Gradient Reversal Layer (GRL) \cite{ganin2015unsupervised} and two Transformer layers which can capture the global dependence among features and improve the accuracy of style classification.
Given a feature map $\mathbf{X\in{\mathbb{R}^{H\times W \times C}}}$ of the feature extractor, the style discriminator can be formulated as:
\begin{equation}
    {D} = \text{Linear}(\text{Transformer}(\text{GRL}(\mathbf{X})))).
\end{equation} \label{discriminator}
Then, we use a generative adversarial training method:
\begin{equation} 
\mathcal{L}_{adv\_G} =   (D(\mathbf{X}_t)) - {l}_s)^2 + (D(\mathbf{Z}_t) - l_s)^2,
\end{equation}
\begin{equation} 
\mathcal{L}_{adv\_D} =   \sum_{d={s,t}} (D(\mathbf{X}_d) - l_d)^2 + (D(\mathbf{Z}_d) - l_d)^2,
\end{equation}
where $\mathcal{L}_{adv\_G}$ represents the training loss for the generator G (\textit{i.e.}, feature extractor) during the adversarial process, while $\mathcal{L}_{adv\_D}$ is the training loss for the discriminator $D$ during the same adversarial process. 
The generator is trained with the least square loss function \cite{mao2017least}, which tries to generate features similar to the source domain and fool the discriminator $D$ from the target domain images, while keeping the discriminator $D$ frozen.
$l_d$ = $0, 1$ denotes that the label comes from the source or target domain, respectively. 
Considering both template feature $\mathbf{Z}_d$ and search feature $\mathbf{X}_d$, the adversarial loss consists of two terms.
Finally, we can obtain the domain-invariant feature by optimizing these two losses alternately.

\begin{table*}[t]
    \centering
     \caption{Comparison of the proposed method with other state-of-the-art trackers on four datasets. The red, green, and blue colors denote the best, the second-best, and the third-best, respectively.}
    \scalebox{0.8}{
    \fontsize{9}{12}\selectfont
    \begin{tabular}{l | c c c | c c c | c c| c c| c}
        \toprule
           &\multicolumn{3}{c}{LSOTB-TIR100}   & \multicolumn{3}{c}{LSOTB-TIR120}  &\multicolumn{2}{c}{PTB-TIR}  &\multicolumn{2}{c}{VTUAV}&VOT-TIR2017\\
           \cline{2-12}
          Tracker&Succ. $\uparrow$&Prce. $\uparrow$ &Norm.Prec. $\uparrow$  &Succ. $\uparrow$ &Prce. $\uparrow$ &Norm.Prec. $\uparrow$ &Succ. $\uparrow$  &Prec. $\uparrow$ &MSR $\uparrow$  &MPR $\uparrow$ &EAO $\uparrow$\\
          \hline
       MLSSNet~\cite{MLSSNet} &40.2 &53.1 &47.7 &45.9 &59.6
       & 54.9 &53.9 &74.1 &- &- &0.286  \\
       
       MMNet~\cite{MMNet} &41.6  &51.7 &47.0 &49.9 &60.9 &56.3  &54.0 &76.0 &- &- &\textcolor{blue}{0.320}  \\

        SiamSRT~\cite{SiamSRT} &51.5 &63.2 &57.8 &55.9 &66.4 & 59.3 &55.4 &75.0 &- &- &-   \\
        
         SiamMask~\cite{wang2019fast} &53.9  &65.9 &59.0 &57.9 &70.5 &63.7 &59.5  &78.0 &39.7 &52.2  &0.258\\ 
        TADT~\cite{TADT} &54.1  &66.5 &58.4 &58.7 &71.0 &63.5 &56.0  &74.0 &- &-  &0.262\\ 
        
      VITAL~\cite{song2018vital} &55.2  &70.6 &63.5 &59.6 &74.8 &68.1 &58.5  &75.0 &- &-  &0.272\\


    SiamRPN++~\cite{li2019siamrpn++} &56.0  &66.1 &60.1 &60.4 &71.1 &64.8 &61.4  &{77.4} &37.6 &47.0  &0.296 \\

    SiamCAR~\cite{guo2020siamcar} &56.6  &68.8 &61.2 &60.1 &72.4 &66.5 &56.6  &71.8 &39.2 &47.5  &0.246   \\

     ATOM~\cite{danelljan2019atom} &56.8  &70.2 &61.3 &59.3 &72.7 &64.5 &61.2  &76.8 &40.6 &49.6  &0.290\\ 
    
    ECO-stir~\cite{zhang2018synthetic} &56.9  &70.2 &61.2 &61.6 &74.9 &67.1 &61.7  &\textcolor{green}{83.0} &- &-  &- \\ 

    PrDiMP~\cite{danelljan2020probabilistic} &59.2  &70.3 &63.1 &62.9 &74.1 &67.2  &57.8 &70.9 &\textcolor{red}{47.5} &\textcolor{red}{56.8}  &-   \\
    
     LTMU~\cite{dai2020high} &59.3  &72.0 &64.4 &62.5 &74.3 &67.3  &- &- &- &-  &-  \\

    TFFT~\cite{TIM2023} &- &- &- &64.0 &77.6 & 70.1 &\textcolor{red}{64.0} &\textcolor{red}{84.2} &- &-  &-  \\ 
    CSWinTT~\cite{song2022transformer} & 60.4 & 72.8& 64.8&64.4 &76.1 &68.5 &57.2  &70.7 &- &-  &-   \\ 
        
    Mixformer~\cite{cui2022mixformer} &-  &- &- &\textcolor{green}{66.0} &\textcolor{blue}{77.8} &\textcolor{blue}{70.3} &61.2  &75.3 &\textcolor{blue}{46.3} &53.7  &-   \\ 
      
    KYS~\cite{bhat2020know} &\textcolor{blue}{60.9}  &\textcolor{blue}{74.0} &\textcolor{blue}{65.8} &64.7 &77.3 &69.8  &60.8 &75.4 &45.4 &53.8  &0.301  \\
         
    DiMP~\cite{bhat2019learning} &\textcolor{green}{61.9}  &\textcolor{green}{74.8} &\textcolor{green}{66.7} &\textcolor{red}{66.2} &\textcolor{red}{78.7} &\textcolor{green}{70.7} &\textcolor{blue}{61.8} &74.9 &46.2 &\textcolor{blue}{54.6}  &\textcolor{red}{0.328}   \\ 
  
         \midrule
          
       PDAT-CAR( \textbf{Ours}) &\textcolor{red}{62.4}  &\textcolor{red}{74.9} &\textcolor{red}{67.0} &\textcolor{blue}{65.7} &\textcolor{green}{78.3} &\textcolor{red}{71.3} &\textcolor{green}{63.0}  &\textcolor{blue}{80.1} &\textcolor{green}{46.6} &\textcolor{green}{54.8}  &\textcolor{green}{0.321}  \\ 
        \bottomrule
    \end{tabular}
    }
   
    \label{all_cp}
\end{table*}

\subsection{Clustering-based subdomain adaptation}
Although the global domain adaptation has aligned the feature distributions of the two domains, we argue that such alignment is insufficient for tracking task. 
Because the global domain alignment treats the data of two domains as a whole, this alignment will cause the feature distribution of different classes of targets to be relatively coupled, which is not conducive to distinguish similar targets in tracking task. 

To solve the above problem, we propose a clustering-based subdomain adaptation network to achieve fine-grained feature alignment. Different from the global domain adaptation, the subdomain adaptation is performed by dividing the data of two domains into multiple similar subcategories, and then performing domain alignment on these subcategories.
Inspired from Local Maximum Mean Discrepancy (LMMD)~\cite{zhu2020deep}, we attempt to use it for subdomain adaptation in our tracking task. 
However, LMMD is designed for classification task, which need the class label of the domain dataset, but our source and target domains lack the class labels information. 
Therefore, we design an online clustering approach to obtain pseudo class labels for our subdomain adaptation network.
Specifically, we first perform a cross-correlation operation on template image feature $\mathbf{Z}_d^{m}$ and search image feature $\mathbf{X}_d^{m}$ from each stage of the feature backbone:
\begin{equation}
\mathbf{F}_{d}^{m} = \mathbf{Z}_d^{m} \star \mathbf{X}_d^{m}, \quad d \in \{s, t\},
\end{equation}
where $m$ is the number of stages in the feature backbone and $d$ denotes the feature comes from source or target domains. 
Then, we use a simple K-means to cluster $\mathbf{F_{d}^{m}}$ into $C$ clusters
which are used as pseudo-labels for our subdomain adaptation.

However, the feature clustering results of each stage are different. 
Which stage of clustering is used as the final class pseudo-label has a significant impact on the tracking results (as shown in Table~\ref{sub_ab} of ablation study).
To this end, we design a simple voting selection mechanism.
After obtaining class pseudo-labels for each stage, we assign different weights to these labels, and the label with the highest voting weight is then selected as final class label for subdomain adaptation.

Subsequently, these labels are input together with the corresponding features into the subdomain adaptation network. 
We utilize the LMMD method for subdomain feature alignment, which needs to calculate two core variables: category weights ${w}$ and kernel functions ${k}$ for both the source and target domains as the following:
\begin{equation}
    {w}^c_i=  \frac{y_{ic}}{\sum_{(\mathbf{x}_j, \mathbf{y}_j) \in S} y_{jc}},
\end{equation}
\begin{equation}
k(\mathbf{X}^s , \mathbf{X}^t ) = \langle \phi(\mathbf{X}^s ), \phi(\mathbf{X}^t) \rangle,
\end{equation}
where $y_{ic}=1$ if a sample $i$ belongs to class $c$, otherwise equals to $0$. $\mathbf{x}_j$ and $\mathbf{y}_j$ denote the $j$-th sample and its corresponding class label in the dataset $S$, while $w^c_i$ is the weight of sample $i$ belonging to class $c$. 
$\mathbf{X}^s $ and $\mathbf{X}^t $ represent the source and target domain features, respectively.
$\phi(\cdot)$ denotes to map the original feature into the Reproducing Kernel Hilbert Space (RKHS) and $\langle \cdot , \cdot \rangle$ is the inner product of vectors. 

Finally, the subdomain adaptation alignment loss can be formulated as:
\begin{small}
\begin{align}
\mathcal{L}_{sub} = \frac{1}{C} \sum_{c=1}^{C} \Bigg[ \sum_{i=1}^{n_s} \sum_{j=1}^{n_s} w_{i}^{sc} w_{j}^{sc} k(\mathbf{X}_{i}^{s}, \mathbf{X}_{j}^{s}) \nonumber \\
+ \sum_{i=1}^{n_t} \sum_{j=1}^{n_t} w_{i}^{tc} w_{j}^{tc} k(\mathbf{X}_{i}^{t}, \mathbf{X}_{j}^{t})  - 2 \sum_{i=1}^{n_s} \sum_{j=1}^{n_t} w_{i}^{sc} w_{j}^{tc} k(\mathbf{X}_{i}^{s}, \mathbf{X}_{j}^{t}) \Bigg],
\end{align}
\end{small}
where $n_s$ and $n_t$ denote the number of samples of the source and target domain datasets.

\subsection{Training and inference} 

\noindent{\textbf{Loss functions.}} The entire domain adaptation TIR tracking framework contains three parts of loss: the global domain adversarial loss ($\mathcal{L}_{adv\_G}$, $\mathcal{L}_{adv\_D}$), the subdomain alignment loss ($\mathcal{L}_{sub}$), and the tracking loss ($\mathcal{L}_{track}$). As shown in Eq.~\ref{trackingloss}, we use the original classification, regression, and centerness losses in the baseline RGB tracker SiamCAR~\cite{guo2020siamcar} as our $\mathcal{L}_{track}$ without any modifications. we integrate the tracking loss with the global domain adversarial loss to form a composite loss, denoted as $\mathcal{L}_{step1}$, while the subdomain alignment loss is treated separately as $\mathcal{L}_{step2}$. 
\begin{align}
\mathcal{L}_{track} = \lambda_{1}\mathcal{L}_{cls} + \lambda_{2}\mathcal{L}_{reg} + \lambda_{3}\mathcal{L}_{cen},
\label{trackingloss}
\end{align}
\begin{align}
\mathcal{L}_{step1} = \mathcal{L}_{adv\_G} + \mathcal{L}_{adv\_D} + \mathcal{L}_{track},
\label{step1}
\end{align}
\begin{align}
\mathcal{L}_{step2} = \mathcal{L}_{sub}.
\label{step2}
\end{align}

\noindent{\textbf{Progressive training strategy.}} Since the subdomain adaptation can be regarded as a fine-tuning for the global domain adaptation, we adopt a progressive training strategy. In each iteration,
we first combine the global domain adaptation loss with the tracking loss and perform backpropagation to obtain the initial global alignment using Eq.~\ref{step1}.
Then we train the subdomain adaptation module to get the subdomain alignment finely using Eq.~\ref{step2}. We believe that this progressive training strategy can achieve better alignment results faster. Because the global alignment can provide a good initialization for subdomain adaptation, the subdomain alignment can accelerate the convergence speed of global domain adaptation.

\noindent{\textbf{Inference.}} In the testing stage, we just use the baseline tracker for TIR tracking. Since the proposed global domain adaptation and subdomain adaptation have learned a good domain-invariant feature, the original tracking method is also adapted to TIR tracking.

\begin{table*}[t]
    \centering
    \caption{Comparison between the proposed method and several state-of-the-art tracker on all attribute subsets of the LSOTB-TIR100. The evaluation metrics include Success rate, Precision, and Normalized precision (S/P/NP). The best results are highlighted in bold font.}
    \scalebox{0.85}{
    \fontsize{9}{12}\selectfont
    \begin{tabular}{c | l || c | c | c |c| c || c }
        \toprule
           Attributes Type & Attributes Name & DiMP & KYS & LTMU & SiamRPN++ & SiamCAR & PDAT-CAR\\
          \hline
 & Deformation & \textbf{61.2}/\textbf{74.6}/\textbf{64.8}  & 57.7/70.8/60.6  & 58.8/71.8/63.5  &53.5/63.4/57.8  & 53.1/65.2/60.1  & 60.2/72.6/63.8
         \\
& Occlusion & \textbf{59.6}/\textbf{72.4}/\textbf{62.8}  & 55.3/67.0/58.0  &54.3/65.7/57.0  & 50.3/59.4/53.7  &51.4/61.6/57.4  &56.8/68.1/60.2
        \\
& Distractor & 54.9/67.2/58.2  & 53.7/65.5/56.6  & 51.5/63.6/55.7  & 49.8/59.1/53.3  & 51.9/63.8/59.5  & \textbf{57.4}/\textbf{69.7}/\textbf{62.6}
        \\ 
& Background clutter & 61.6/73.8/66.0  & 62.2/74.8/66.7  & 57.3/68.9/62.3  & 56.0/64.9/59.7  & 55.5/66.7/63.4  &\textbf{62.6}/\textbf{75.0}/\textbf{67.8}
       \\ 
& Out of view & 60.2/65.5/62.7  & 62.7/68.6/66.1  & 58.5/67.2/61.2  & 60.3/69.5/65.9  & 59.3/69.3/68.1  &\textbf{64.5}/\textbf{73.5}/\textbf{70.7}
      \\ 
Challenge & Scale variation & 67.8/79.4/74.8  & 69.2/81.8/77.1  & 67.2/80.8/76.3  & 64.2/76.1/71.6  & 66.2/80.5/78.6  &\textbf{72.1}/\textbf{86.8}/\textbf{82.6} 
     \\ 
& Fast motion & \textbf{66.8}/76.8/73.3  & 64.2/74.6/71.6  & 65.5/\textbf{77.5}/73.3  & 61.8/71.7/67.6  & 64.9/76.6/\textbf{75.4}  &66.6/76.8/73.3 
       \\ 
& Motion blur & 65.1/78.1/70.5  & 66.1/80.4/72.6  & \textbf{66.2}/\textbf{82.1}/71.6  & 58.1/70.4/62.5  & 58.1/71.7/66.6  &65.0/78.9/\textbf{73.7}  
     \\ 
&Thermal crossover  & 46.3/64.5/46.0  & 49.6/70.2/47.9  & 46.1/61.2/50.1  & 47.3/61.3/50.8  & 41.6/56.4/45.3  &\textbf{56.4}/\textbf{73.7}/\textbf{57.6}
    \\ 
& Intensity variation & 76.6/85.6/81.1  & 79.2/88.8/82.8  & 71.1/84.5/80.1  & 76.5/84.4/80.1  & \textbf{80.9}/91.5/\textbf{85.5}  &77.4/\textbf{91.7}/84.6
    \\
& Low resolution & 67.4/88.3/73.7  & \textbf{71.1}/\textbf{94.8}/77.3  & 65.1/82.4/72.3  & 60.1/75.6/67.7  & 65.3/83.9/\textbf{78.8}  &66.7/86.6/75.6
     \\
& Aspect ratio variation & 63.4/77.6/69.2  & \textbf{64.1}/\textbf{79.0}/\textbf{70.5}  & 60.4/73.1/66.3  & 57.8/69.4/62.5  & 58.4/70.4/67.1  &63.8/74.6/69.6
    \\ 
    \hline
& Vehicle-mounted & 72.7/69.6/65.7  & 72.8/82.7/78.5  & 68.5/82.0/75.2  & 66.8/78.1/74.6  & 75.9/90.7/\textbf{87.5}  &\textbf{76.4}/\textbf{91.2}/ 85.7 
     \\ 
Scenario& Drone-mounted & 56.0/68.8/63.3  & 54.2/67.9/60.3  & 53.3/64.6/60.8 & 53.7/64.2/59.4  & 54.7/68.4/65.5  &\textbf{63.2}/\textbf{76.4}/\textbf{71.0}
     \\
& Surveillance & \textbf{60.5}/\textbf{73.1}/\textbf{63.4}  & 56.1/67.0/58.0  & 52.4/63.6/57.3  & 51.2/59.4/55.1  & 50.6/59.6/56.8  &54.9/66.7/60.5
     \\ 
& Hand-held & \textbf{66.9}/\textbf{81.3}/\textbf{70.2}  & 65.6/80.8/69.5  & 65.6/79.7/67.7  & 58.3/68.9/60.1  & 57.6/70.3/64.1  & 63.8/76.0/68.7
   \\
   \hline
All & All & 61.9/74.8/66.7  & 60.9/74.0/65.8  & 59.3/72.0/64.4  & 56.0/66.1/60.1  & 56.6/68.8/61.2  &\textbf{62.4}/\textbf{74.9}/\textbf{67.0}
       \\ 
        \bottomrule
    \end{tabular}
    }
    \label{lsotb100_att_cp}
\end{table*}

\section{Experiment}
\subsection{Implementation details}
We train the proposed method on $4$ NVIDIA RTX A4000 GPUs using PyTorch framework.
We choose the RGB tracker SiamCAR \cite{guo2020siamcar} as the baseline method. The discriminator is optimized using the Adam optimizer and we adopt a poly learning rate policy with a power of 0.8 to decay the base learning rate of 0.005. 
We train the proposed method with 20 epochs and set batchsize to $24$. 
we use a pre-trained SiamCAR as initialization and the used RGB datasets are the same as this paper. 
We conduct the global domain adaptation on all $4$ stage feature blocks and only on last stage for the subdomain adaptation. 
The feature backbone is ResNet50.
We set $2$ to $10$ clusters and then use Silhouette criterion to adaptive select optimal number in the cluttering-based subdomain adaptation.
The weights in label selection mechanism are set $1$ to $4$  for stage 1 to stage 4 equidistantly.

\begin{table}[!t]
    \centering
     \caption{Comparison of the proposed method PDAT-CAR with the baseline tracker SiamCAR. $\Delta$ represents the absolute gain by PDAT.}
    \scalebox{0.85}{
    \fontsize{10}{13}\selectfont
    \begin{tabular}{cccc}
        \toprule
        &\multicolumn{2}{c}{LSOTB-TIR100} \\
        Trackers & Prec.($\uparrow$) & Norm.Prec($\uparrow$) & Succ.($\uparrow$)\\
        \hline
        SiamCAR   & 68.8 & 61.2 & 56.6  \\ 
        PDAT-CAR  & 74.9 & 67.0 & 62.4 \\
        $\Delta$\text{CAR} & +6.1 & +5.8 & +5.8 \\
        \hline
        \hline
        &\multicolumn{2}{c}{LSOTB-TIR120} \\
        Trackers & Prec.($\uparrow$) & Norm.Prec($\uparrow$) & Succ.($\uparrow$)\\
        \hline
        SiamCAR   & 72.4 & 66.5 & 60.1 \\ 
        PDAT-CAR  & 78.3 & 71.3 & 65.7 \\
        $\Delta$\text{CAR} & +5.9 & +4.8 & +5.6\\
        \hline
        \hline
         &\multicolumn{2}{c}{ PTB-TIR } \\
        Trackers & Prec.($\uparrow$) & Norm.Prec($\uparrow$) & Succ.($\uparrow$)\\
        \hline
        SiamCAR   & 71.8 & $-$ & 56.6 \\ 
        PDAT-CAR  & 80.1 & $-$ & 63.0 \\
        $\Delta$\text{CAR} & +8.3 & $-$ & +6.4 \\
        \bottomrule
    \end{tabular}
    }
    \label{CAR_cp}
\end{table}

\subsection{ Datasets and evaluation metrics}
 We evaluate the proposed method on four widely used TIR tracking datasets, including LSOTB-TIR100~\cite{liu2023lsotb}, LSOTB-TIR120~\cite{LSOTBTIR120}, PTB-TIR~\cite{liu2019ptb}, and VOT-TIR2017 \cite{8265440}, and VTUAV~\cite{VT-UAV}. 
 The LSOTB-TIR100 and LSOTB-TIR120 are two large-scale and high-diversity TIR tracking benchmarks, consisting of 100 and 120 evaluation sequences, respectively.  
 They use precision, normalized precision, and success rate as evaluation metrics. 
 The PTB-TIR dataset is designed for evaluating TIR pedestrian trackers, which has $60$ sequences and also uses precision and success rate as evaluation metrics. 
 The VOT-TIR2017 dataset contains $25$ TIR sequences, and often use Expected Average Overlap (EAO) to evaluate the accuracy and robustness of a tracker. 
 The VTUAV dataset is comprised of multi-modality data for Unmanned Aerial Vehicle (UAV) tracking, including a total of 176 test sequences across both RGB (Red-Green-Blue) and Thermal Infrared (TIR) modalities. For our evaluation purposes, we have exclusively utilized the TIR sequences, employing the Maximum Success Rate (MSR) and Maximum Precision Rate (MPR) as the key performance metrics.

\begin{figure}[t]
\centering
\scalebox{0.40}{
\includegraphics{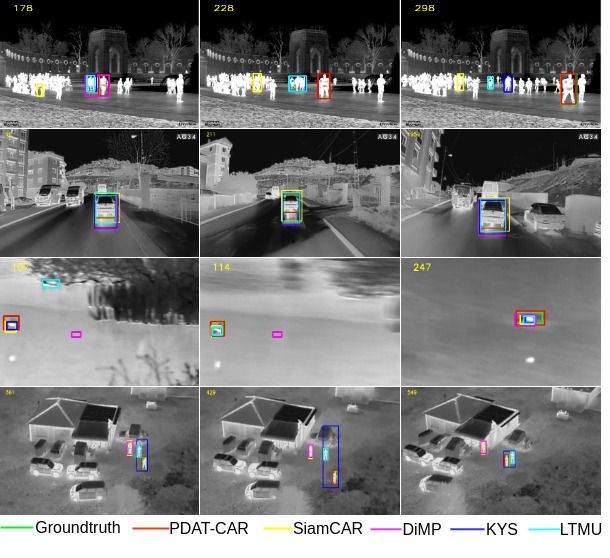}
}
\caption{Qualitative comparison between the proposed method (PDAT-CAR) and several state-of-the-art trackers on the similar distractor and background clutter challenges of LSOTB-TIR100.
}
\label{track_vis}
\end{figure}

\subsection{Comparison with state-of-the-arts} 
To comprehensive evaluate the performance of the proposed method PDAT-CAR in TIR tracking, we first compare it with the original baseline on three datasets, as shown in Table~\ref{CAR_cp}. The results show that the proposed method has significantly improved on all the three datasets, achieving an absolute gain of over 6\% and 5\% in precision and success rate, respectively.
In addition, we conduct extensive comparison with other top-performance trackers on five datasets, the results are shown in below.

\begin{figure}[t]
\centering
\scalebox{0.33}{
\includegraphics{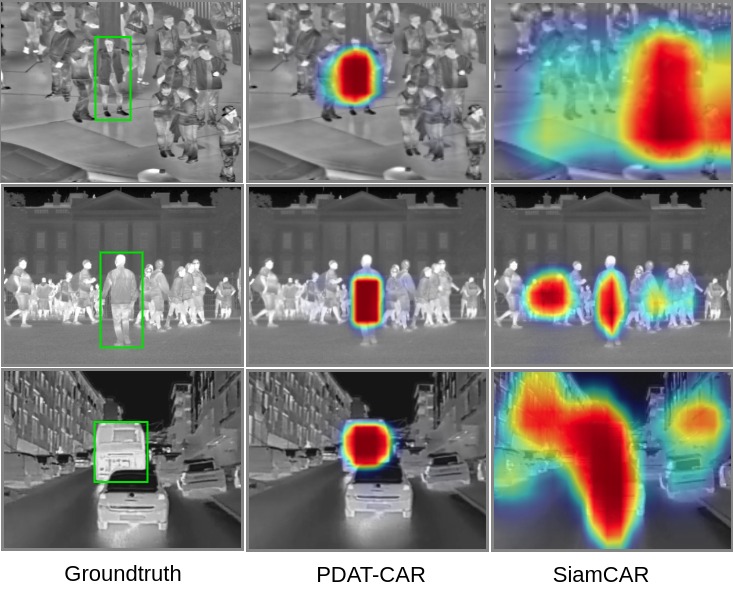}
}
\caption{Comparison of the confidence maps generated by the proposed PDAT-CAR and the baseline SiamCAR on several challenging sequences of LSOTB-TIR120. The green bounding box represents the groundtruth of the target. 
PDAT-CAR significantly reduces interference from the background clutter and similar distractor.
}
\label{heatmap}
\end{figure}

\noindent{\textbf{Results on LSOTB-TIR100.}} As shown in Table~\ref{all_cp}, the proposed method achieves best success rate and precision. 
Compared with Siamese series method, e.g.,  SiamRPN++, SiamMask, and SiamCAR, the proposed method has an improvement of more than 5\% in success rate. This shows that our domain adaptation framework can indeed transfer the useful knowledge learned in RGB datasets to TIR tracking effectively.
Although online learning series method can effectively learn features in TIR domain,, e.g., VITAL, ATOM, and DiMP, our method also shows superiority on all three metrics. This shows that the learned domain-invariant feature can be well adapted to TIR tracking. 
Figure~\ref{track_vis} shows that our proposed method PDAT-CAR gets more precise tracking results compared to the baseline and other state-of-the-art (SOTA) trackers.
When there are many similar distractors and background clutters, the proposed method can distinguish the target and obtains more accurate tracking results. This demonstrate that the proposed domain adaptation method can achieve fine-grained features transfer.

\begin{figure}[t]
    \centering
\scalebox{0.26}{
        \includegraphics{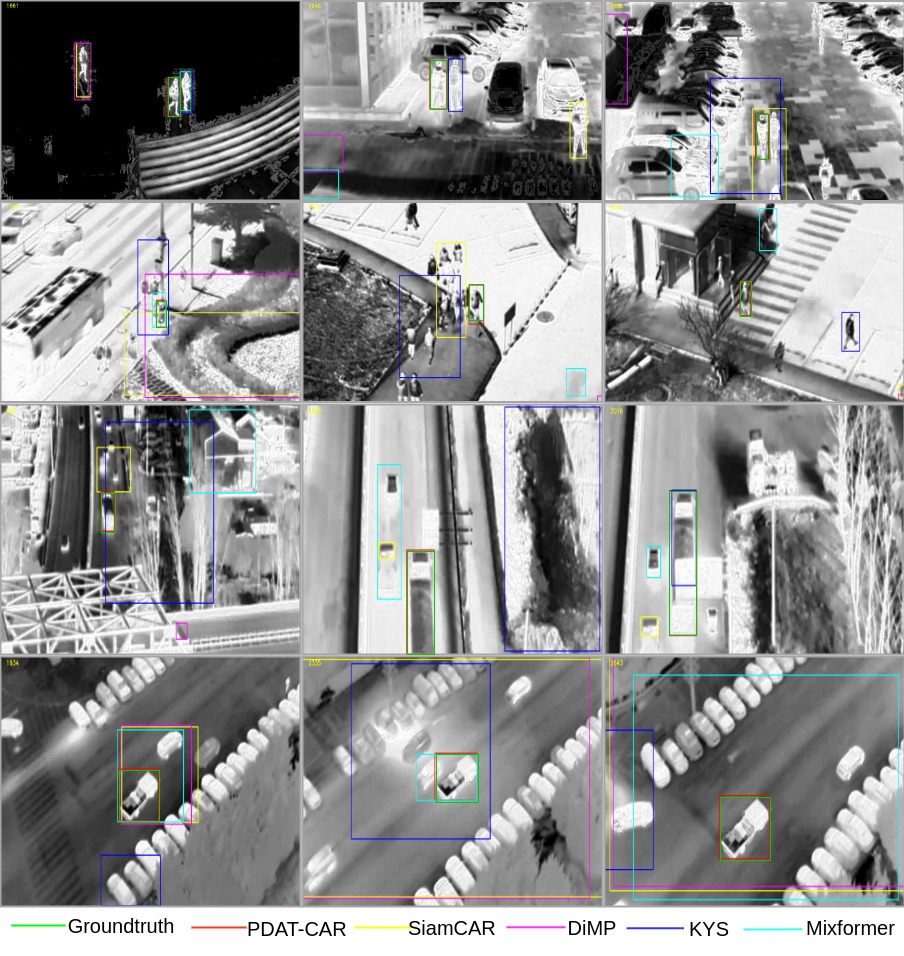}
        }
    \caption{Compared the proposed PDAT-CAR with other state-of-the-art trackers on the VTUAV dataset.}  \label{VTUAV}
\end{figure}

\begin{figure*}[t]
    \centering
\scalebox{0.42}{
        \includegraphics{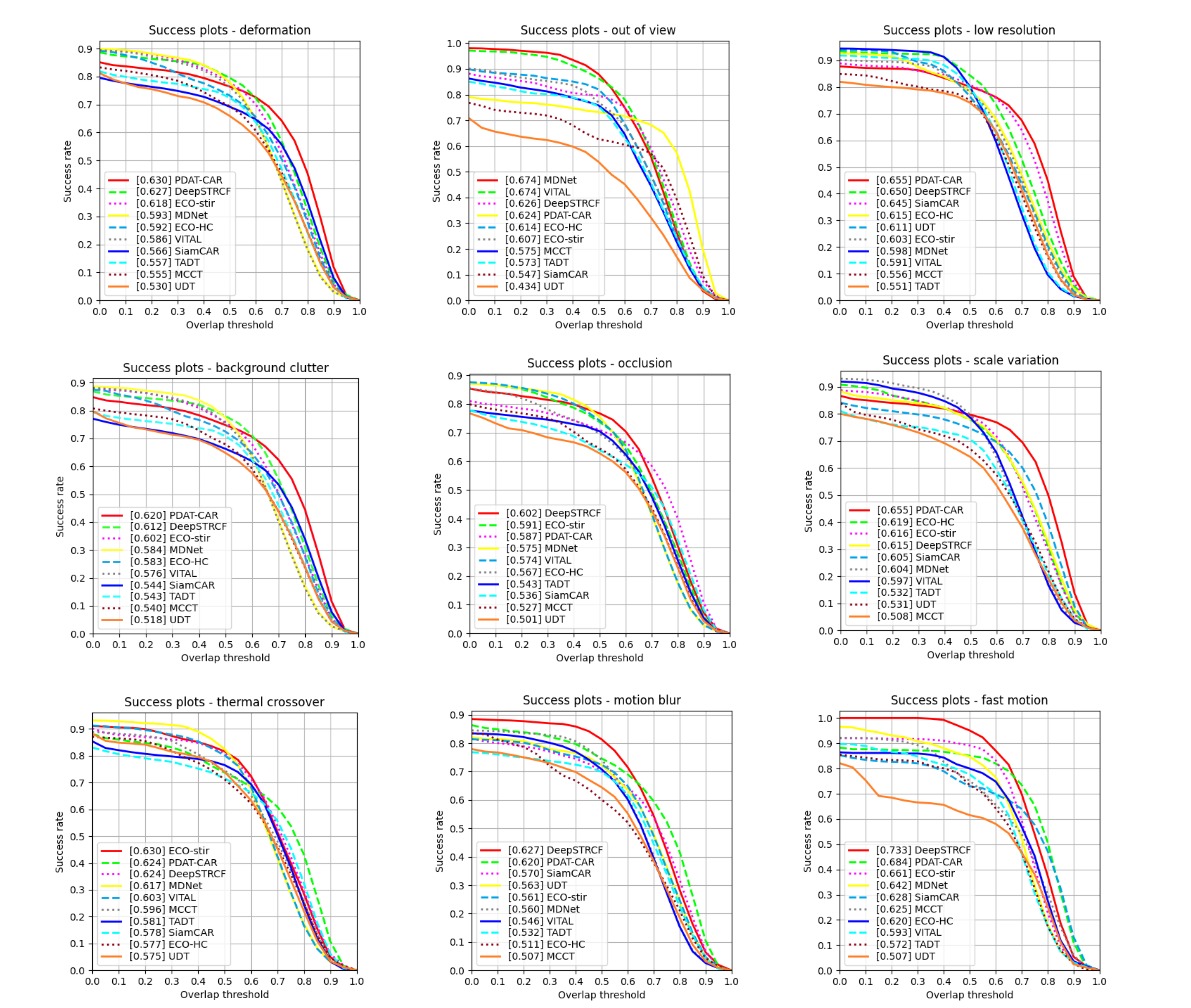}
        }
    \caption{Comparison between the proposed method and state-of-the-art trackers on all challenge subset of the PTB-TIR dataset.}  
    \label{PTBTIR}
\end{figure*}

\noindent{\textbf{Results on LSOTB-TIR120.}} 
As shown in Table~\ref{all_cp}, our method also achieves the best normalized precision and the second-best success rate.
Compared with Transformer based methods, e.g., CSWinTT and Mixformer, the proposed method achieves comparable performance, although our feature network only uses the smaller ResNet50.
This shows that the domain shift problem has a huge impact on tracking performance.
We utilize a Grad-CAM \cite{selvaraju2017grad} to visualize the confidence maps of PDAT-CAR and the baseline method SiamCAR, as shown in Figure~\ref{heatmap}. 
It shows that the original tracking method SiamCAR is easily interfered by similar targets or background, while the proposed method can accurately identify and locate the target.

\noindent{\textbf{Results on PTB-TIR.}} As shown in Table~\ref{all_cp}, we can see that our method also achieves the second-bes success rate. Although ECO-stir shows excellent performance on the PTB-TIR dataset, its performance is average on other datasets.  Figure~\ref{PTBTIR} presents a comparison of several state-of-the-art trackers on the pedestrian dataset PTB-TIR in terms of attributes. It can be observed that our proposed PDAT-CAR shows significant improvements over the baseline SiamCAR, particularly in deformation(+6.4\%) and background clutter(+7.6\%). Not only does it perform excellently on the LSOTB-TIR , but it also excels on the pedestrian dataset PTB-TIR, demonstrating the generalizability and good domain transfer capability of our method.

\begin{figure*}[t]
    \centering
\scalebox{0.45}{
        \includegraphics{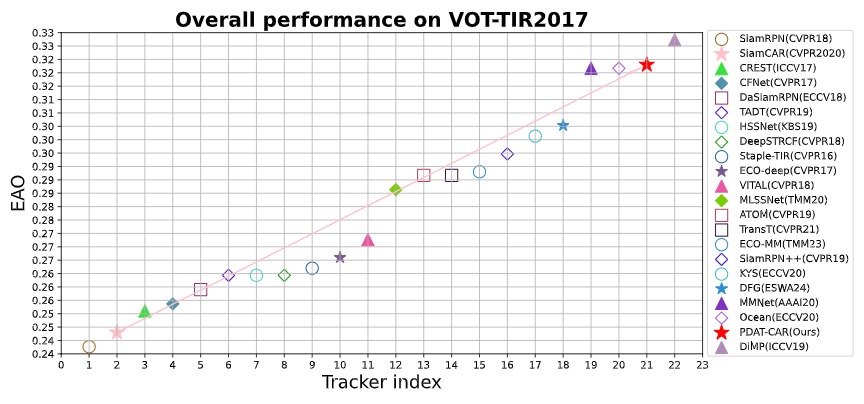}
        }
    \caption{Compared the proposed PDAT-CAR with other state-of-the-art trackers on the VOT-TIR2017 dataset.}  \label{EAO}
\end{figure*}

\noindent{\textbf{Results on VOT-TIR2017.}} As shown in Table~\ref{all_cp}, our method gets the second-best EAO. Different from MLSSNet and MMNet which have good performances on VOT-TIR2017 but has poor performances on LSOTB-TIR100, our method obtains favorable results on both two datasets. This shows that our method achieves good robustness for different scenarios. Figure~\ref{EAO} shows that our method achieves significant enhancements compared to the benchmark SiamCAR. In the figure, {\color{pink} \faStar} indicates the baseline SiamCAR, and {\color{red} \faStar} signifies our approach, PDAT-CAR. It is observable that our method elevates the benchmark from the second-to-last score to the second-highest score, surpassing numerous state-of-the-art trackers. This further demonstrates that the proposed domain adaptation method obtains fine-grained feature transfer.

\noindent \textbf{Results on VTUAV.} 
As shown in Table~\ref{all_cp}, our method obtains the second-best performance in both MSR and MPR, with a 7.4\% improvement in MSR and a 7.3\% improvement in MPR over the baseline, significantly enhancing the recognition capabilities for small target objects.  
Although PrDiMP achieves the best results, its performance is not as satisfactory when evaluated on other datasets. Conversely, our method demonstrates outstanding performance across all datasets. At the same time, Figure~\ref{VTUAV} shows that PDAT-CAR obtains good performance and significantly enhances the recognition capabilities for small target objects. This indicates that our method achieves excellent domain transfer efficacy and robustness for different scenarios.

\subsection{Ablation study}
To demonstrate the effectiveness of each components of the proposed method, we conduct an ablation study for each module of PDAT on the LSOTB-TIR100 and PTB-TIR benchmarks. 

\noindent{\textbf{Network architectures.}} We first validate each part of the network architecture of the PDAT, the results are shown in Table~\ref{overall_ab}.
The first blank row represents the result of the original baseline SiamCAR and the results in parentheses in each row below represent improvements relative to the baseline. 
From the second row results, we can see that when we only use SAM-based data preprocessing to get TIR training pairs of pseudo label for training of the baseline, the success rate increases by only 2.2\%. 
While we add the global domain adaptation module (AGDA) and subdomain adaptation module (CSDA), the success rate increases by 3.6\% and 2.9\%, respectively.
This demonstrates that the proposed the global domain adaptation and subdomain adaptation module  are effective.
From the fourth and fifth rows results, we can see that when we use the designed label section mechanism in the class pseudo-label generation, the subdomain adaptation has been further improved (+2.9\% $\rightarrow$ +3.8\%).
This shows that more accurate class labels can enhance the performance of subdomain alignment.
From the last row results, we can see that when we use AGDA and CSDA together with the progressive training, the success rate gets a larger boosting (+3.6\% $\rightarrow$ +5.8\%; +3.8\% $\rightarrow$ +5.8\%) than using either of them alone.
This demonstrates that the proposed progressive domain adaptation can obtain the better feature alignment and transfer effect.

\begin{table}[!t]
    \centering
     \caption{ Ablation study of the label selection mechanism in the clustering-based subdomain adaptation on LSOTB-TIR100. Clustering Label and Align denote which stage class pseudo label comes from and which stage feature of subdomain alignment used, respectively.
    }
    \scalebox{0.8}{
    \fontsize{10}{13}\selectfont
    \begin{tabular}{ccccc}
        \toprule
         & & Prec. & Norm.Prec. & Succ. \\
         \midrule
         Align  &Clustering Label & & & \\
         \midrule
          \multicolumn{2}{c}{SAM-only}  & 70.3 & 61.8 &58.8 \\ 
          Stage1-4  & Stage1-4  & 70.6(+0.3) & 61.9(+0.1) &58.9(+0.1)  \\ 
         Stage4 & Stage4 & 71.4(+1.1) & 63.1(+1.3) &59.5(+0.7)  \\ 
          Stage1-4 & Stage4 & 69.7(-0.6) & 61.4(-0.4) &58.3(-0.5)  \\ 
         Stage4 & Weighted Stage1-4  & 72.3(+2.0) & 64.9(+3.1) &60.4(+1.6)  \\ 
        \bottomrule
    \end{tabular}
    }

    \label{sub_ab}
\end{table}

\noindent{\textbf{Label selection mechanism. }} Second, we validate the label selection and alignment way of the subdomain adaptation as shown in Table~\ref{sub_ab}. SAM-only is used as the baseline for comparison, which is the same as the second row of Table~\ref{overall_ab}.
The second row results show that when we use class pseudo label and align the subdomain on all stages, the results are basically the same as the baseline method.
While we use the stage4 for clustering and aligning, the results were slightly improved. 
These results show that only clustering or aligning on the last stage, the aligned effect of the subdomain is better.
However, it is not sure that whether the alignment or better clustering labels are working.
Therefore, we fix the clustering at stage4 and then align the subdomain on all stages, as shown in fourth row of Table~\ref{sub_ab}.
The results find that the performance dropped significantly. This demonstrate that the alignment on last stage is critical for subdomain adaptation in our framework.
Finally, we fix the alignment on stage4 and use a simple weighted voting mechanism on pseudo class label selection of stage1 to stage4. The results show that the success rate has a significantly improved (+0.7\% $\rightarrow$ +1.6\%) compared to use only the last stage of pseudo label.
This demonstrates that the proposed label selection mechanism can obtain more accurate pseudo label for fine-grained subdomain adaptation.

\begin{table}[t]
    \centering
     \caption{Ablation study of the network architecture of PDAT on LSOTB-TIR100. 
    SAM, AGDA, CSDA, and LS denote the data preprocessing with SAM, adversarial-based global domain adaptation, clustering-based subdomain adaptation, and label selection mechanism, respectively.}
    \scalebox{0.78}{
    \fontsize{10}{13}\selectfont
    \begin{tabular}{cccccccc}
        \toprule
        SAM & AGDA & CSDA & LS  && Prec. & Norm.Prec & Succ.\\
      \midrule
        &&&&&68.8 &61.2 &56.6\\
        \midrule
        \checkmark   &&&&& 70.3(+1.5) & 61.8(+0.6) & 58.8(+2.2)  \\ 
        
        \checkmark  &\checkmark &&&& 71.5(+2.7) & 65.2(+4.0) & 60.2(+3.6)\\
        
        \checkmark  &&\checkmark &&&71.4(+2.6) & 63.1(+1.9) & 59.5(+2.9)\\

        \checkmark  &&\checkmark &\checkmark && 72.3(+3.5) & 64.9(+3.7) & 60.4(+3.8)\\
        \midrule
         \checkmark &\checkmark &\checkmark &\checkmark && 74.9(+6.1) & 67.0(+5.8) & 62.4(+5.8)\\
        \bottomrule
    \end{tabular}
    }
    \label{overall_ab}
\end{table}

\noindent \textbf{Preprocessing methods.} 
Third, we compare three kinds of preprocessing methods on two benchmarks, as shown in Table~\ref{ablationP}. 
We can see that the SAM-based preprocesing method gets the best performance on both two datasets.
We attribute the SAM model can segment more diverse and accurate pesudo-label training samples, which is important to train the proposed domain adaptation tracking framework.
Compared with the recognition-based method, the SAM-based method gains a $3.6\%$ and $4.2\%$ success rate on the two datasets, respectively.
This is because the recognition-based method generates the fewest pesudo training samples 1.4M which is not enough to train the proposed framework well.
What's more, we can see that the detection based method achieves an obvious improvement on both two datasets compared with the recognition-based method. This is because the detection model can obtain more training samples with more accurate pesudo-labels. 
These results show that more training samples and more accurate labels can significantly improve the transfer effect and tracking performance of our method.

\begin{table}[t]
\small
\centering
\caption{Comparison of the three proposed preprocessing methods on two benchmarks.} 
 \fontsize{8}{10}\selectfont
\begin{tabular}{c|c|ccc|cc}
\toprule
 \multirow{2}{*}{Method} & \multirow{2}{*}{\tabincell{c}{Num. of \\\ Samples}} & \multicolumn{3}{c|}{LSOTB-TIR100}  & \multicolumn{2}{c}{PTB-TIR} \\
\cline{3-7}
   & & Suc.    & Norm.Pre.   & Pre. & Suc.  & Pre.  \\
\midrule
Recog. based & 1.4M & 58.8 &63.1 &70.9 &58.8&74.2\\
Det. based & 2.3M & 61.1 &65.2 &72.6  &61.6&77.4\\
SAM based & 10.2M & 62.4 &67.0 &74.9  &63.0 &80.1\\
\bottomrule
\end{tabular}
\label{ablationP}
\end{table}

\section{Conclusion}
\label{conclusion}
In this paper, we propose a progressive domain adaptation TIR tracking framework which transfers the useful knowledge learned from large-scale labeled RGB datasets to TIR tracking effectively. 
The proposed framework learns domain-invariant features by aligning the distributions of TIR and RGB domains using an adversarial-based global domain adaptation and a clustering-based subdomain adaptation.
The proposed global domain adaptation and subdomain adaptation align the distribution of the two domains from coarse to fine through a progressive training strategy, which can achieve fine-grained feature transfer to boost TIR tracking.
To train the proposed framework, we also collect one of the largest unlabeled TIR datasets to date, which is critical for unsupervised TIR tracking.

\ifCLASSOPTIONcaptionsoff
  \newpage
\fi





\bibliographystyle{IEEEtran}
\bibliography{IEEEabrv,Bibliography}

\vfill


\end{document}